\documentclass[lettersize,journal]{IEEEtran}

\usepackage[dvipsnames]{xcolor}
\usepackage{amsmath,amsfonts}
\usepackage{algorithmic}
\usepackage{array}
\usepackage[caption=false,font=normalsize,labelfont=sf,textfont=sf]{subfig}
\usepackage{textcomp}
\usepackage{stfloats}
\usepackage{url}
\usepackage{verbatim}
\usepackage{graphicx}
\usepackage{balance}
\usepackage[colorlinks=true, linkcolor=blue, urlcolor=blue]{hyperref}
\usepackage{pgfplots}
\pgfplotsset{compat=1.18} 
\usepackage{booktabs}
\usepackage{pifont}
\usepackage[ruled,linesnumbered]{algorithm2e}
\usepackage{color, colortbl}
\usepackage{makecell}
\usepackage{soul}
\usepackage{todonotes}

\begin{document}

\newcommand{\fabiocomment}[1]{\todo[color=purple!20, inline, author=Fabio]{#1}}
\newcommand{\davidecomment}[1]{\todo[color=blue!20, inline, author=Davide]{#1}}
\newcommand{\mohamedcomment}[1]{\todo[color=green!20, inline, author=Mohamed]{#1}}

\newcommand{\fabio}[1]{\textbf{\textcolor{purple!75}{#1}}}
\newcommand{\davide}[1]{\textbf{\textcolor{blue!75}{#1}}}
\newcommand{\mohamed}[1]{\textbf{\textcolor{green!75}{#1}}}

\newcommand{\warning}[1]{\textbf{\color{red!90}{#1}}}

\definecolor{ForestGreen}{RGB}{34,139,34}

\newcommand{\daca}{SF-DACA}

\newcommand{\impp}[1]{{\textcolor{Green}{+#1}}}
\newcommand{\impn}[1]{{\textcolor{BrickRed}{-#1}}}

\title{Leveraging Confident Image Regions for Source-Free Domain-Adaptive Object Detection}

\author{
Mohamed L. Mekhalfi, 
Davide Boscaini,
Fabio Poiesi

\thanks{
Corresponding author: M. L. Mekhalfi (email: mmekhalfi@fbk.eu).
M. L. Mekhalfi, D. Boscaini, F. Poiesi are with Fondazione Bruno Kessler (FBK), 38123 Trento, Italy.
This paper is supported by European Union’s Horizon Europe research and innovation programme under grant
agreement No 101092043, project AGILEHAND (Smart Grading, Handling and Packaging Solutions for Soft and Deformable Products in Agile and Reconfigurable Lines).
}
}


\maketitle

\begin{abstract}
Source-free domain-adaptive object detection is an interesting but scarcely addressed topic.
It aims at adapting a source-pretrained detector to a distinct target domain without resorting to source data during adaptation.
So far, there is no data augmentation scheme 
tailored to source-free domain-adaptive object detection.
To this end, this paper presents a novel data augmentation approach that cuts out target image regions where the detector is confident, augments them along with their respective pseudo-labels, and joins them into a challenging target image to adapt the detector.
As the source data is out of reach during adaptation, we implement our approach within a teacher-student learning paradigm to ensure that the model does not collapse during the adaptation procedure.
We evaluated our approach on three adaptation benchmarks of traffic scenes, scoring new state-of-the-art on two of them. 
\end{abstract}

\section{Introduction}\label{introduction}
Object detection is a hot and challenging computer vision topic that benefits many vision tasks~\cite{lu2023cross, roy2022fast, feng2021review, zhang2020multi}. Its challenge stems from several factors such as occlusion, background clutter, and disparity of object appearance~\cite{diwan2023object}. 
It has animated an exponential growth in the scientific landscape over the past two decades~\cite{zou2023object}.
Source-free domain-adaptive object detection is a research topic that aims at adapting an object detector pretrained on a source dataset to a target dataset drawn from a disjoint distribution, without human labeling.
It is relevant in traffic and transportation scenes that are often characterized with dense crowds and vehicles~\cite{xiong2021source, li2022source}.

The progress witnessed in object detection is owed mainly to the advent of deep learning methodologies and the accessibility to pretrained models~\cite{krizhevsky2012imagenet, simonyan2014very} via transfer learning and opportune fine-tuning, which borrows the knowledge gained by a deep model already pretrained (typically on a large dataset) on an auxiliary task (e.g., image classification) and tailors it to another related task (e.g., object detection)~\cite{zhu2023transfer, kora2022transfer}.
Ideally, this would require target ground-truth for optimal transferability. In real scenarios, however, this is not necessarily guaranteed, which may constitute generalization bottlenecks for data-hungry deep learning paradigms.
In this respect, domain adaptation (DA) has emerged as a potential remedy~\cite{oza2023unsupervised, li2023domain, li2021free}. Essentially, it aims at leveraging the knowledge learned by a model on annotated source data (also referred to as source domain/distribution), in order to adapt the model to a target domain that lacks or has partial annotation (i.e., unsupervised, few-shot, or semi-supervised)~\cite{oza2023unsupervised}.
Unlike the fairly straightforward transfer learning, DA typically requires building specific methodologies for the adaptation problem at hand. Precisely, DA comes at the expense of distribution mismatch across domains (source vs target), which can be due to weather changes~\cite{weather, weather2}, acquisition time (e.g., daytime vs nighttime)~\cite{daynight}, nature of data (e.g., synthetic vs real)~\cite{synthetic}, locations (e.g., different cities)~\cite{city}, and camera (e.g., RGB vs thermal)~\cite{thermal}. 


Relevant research on DA for object detection suggests several settings.
Unsupervised DA (UDA) for object detection represents the majority of contributions~\cite{oza2023unsupervised}. It proceeds by pretraining a base detector on a source domain, and then adapts it on the target images involving the source data again to maintain the model's knowledge gained on the source and prevent eventual collapse (i.e., catastrophic forgetting \cite{aleixo2023catastrophic}).
However, real-life cases imply different situations because source data may not be granted due to (for instance) privacy concerns~\cite{chen2023exploiting, hao2024simplifying}.
This gave rise to an extreme variant of UDA, namely source-free UDA (SF-UDA), where only the target domain images are accessible at the adaptation phase of the already source-pretrained detector. 
Hence, it poses major challenges due to the lack of source groundtruth.

Compared to UDA, SF-UDA is far less addressed in the relevant literature, and there is plenty of room for improvement. Data augmentation paradigms have been shown to be useful in image classification, object detection and segmentation purposes~\cite{fang2024data, bosquet2023full, xu2023comprehensive, kumar2024image, goceri2023medical}. They offer the advantage of manipulating the input images/labels without altering the model's architecture. Although data augmentation was addressed (at least scarcely) in the UDA setting, to our awareness it was not leveraged in the SF-UDA scenario. To this end, this paper devises a novel data-augmentation based approach for SF-UDA, namely SF-DACA which stands for \textit{source-free detect, augment, compose, adapt}, indicating four sequential steps that the adaptation process undergoes. SF-DACA capitalizes on the knowledge learned from the source domain during the pretraining phase to derive high quality predictions from the target domain, and further treat them as pseudo-labels to increment the knowledge on the target domain. This is achieved by selecting confident target image regions and their respective pseudo-labels and joining them via augmentations into one image to self-train the detector.
Because the detector does not possess knowledge on the target, pseudo-labels often comprise false positives, which if augmented as explained earlier, will grow and mislead the adaptation process if the knowledge on the source domain is not maintained, ultimately causing a collapse as evidenced experimentally in Sec. \ref{experiments}. To cope with this, a teacher-student approach is proposed in this paper, where both the teacher and the student models are initially pretrained on the source domain. During adaptation, the dynamic student is adapted towards the target via the four aforementioned steps. Simultaneously, the student's knowledge is maintained through a consistency loss against the predictions of the static teacher on the target images. We increment the teacher knowledge on the target by partially inheriting student weights using exponential moving average \cite{ema}.          

This paper is an extension of our prior work, DACA~\cite{mekhalfi2023daca}, in several aspects.
First, we surveyed more related works, and we summarized a comprehensive table that reports key-features, and the setup of different works.
Second, we extended DACA~\cite{mekhalfi2023daca} from UDA to the SF-UDA setting.
To our knowledge, SF-DACA is the first approach that exploits data augmentation in the SF-UDA scenario. 
Third, we extended the experimental validation and we achieved superior results w.r.t to prior works.

The rest of the paper is organized as follows.
Sec.~\ref{relatedworks} reviews existing relevant art.
Sec.~\ref{method} describes the proposed SF-DACA.
Sec.~\ref{experiments} conducts the experiments and their discussions.
Sec.~\ref{conclusion} wraps up the paper.

\section{Related works}\label{relatedworks}

\subsection{Object detection}

Existing works can be categorized into two-stage or one-stage detectors.
Regarding the former one, regions with CNN (R-CNN)features~\cite{girshick2014rich} introduced impressive performance.
Fast R-CNN~\cite{Fast} passes the entire image through the CNN once, making inference much faster.
Faster R-CNN~\cite{faster} integrates region proposal networks to predict region proposals that require only
negligible additional computation.

As per the one-stage detectors, they seek to abandon the proposals step, and directly predict the bounding boxes and class labels for objects in one pass.
For instance, single shot multibox detector (SSD)~\cite{ssd} produces a fixed set of default bounding boxes and predicts the class label and the bounding box coordinates for each one. 
You Only Look Once (YOLO)~\cite{yolo} divides the image into an even grid, where each grid is responsible for detecting objects whose center falls within it, drastically reducing the processing overheads.
YOLO has evolved into several versions, a chronological narrative was conducted in~\cite{sapkota2024yolov10}. 
RetinaNet~\cite{focal} addresses the class imbalance via a Focal Loss that focuses training on a sparse set of hard examples and prevents the vast number of easy samples from overwhelming the detector~\cite{focal}.
CenterNet~\cite{zhou2019objects} assumes that each pixel in a predicted heatmap stands for a keypoint that represents the center of a potential object in the input image, and regresses to all other object properties.
In~\cite{detr}, an encoder-decoder detection transformer that predicts all objects at once was presented. 
EfficientDet \cite{tan2020efficientdet} combines EfficientNet backbone for feature extraction, and a bi-directional feature pyramid network for efficient feature fusion.

\vspace{-10pt}
\subsection{Unsupervised domain adaptation for object detection}

UDA setting is the most addressed in the literature.
Various schemes have emerged so far, such as style transfer, feature alignment, and self-training (see the first part of Tab.~\ref{tab:sota}).

\textbf{Style transfer} methods attempt to mitigate the domain shift at image level by modifying the appearance of source images to make them similar to target images, thus improving the generalization of an object detector on the target domain.
Rodriguez et al.~\cite{rodriguez2019domain} adopt a feed-forward stylization method proposed in~\cite{huang2017arbitrary}. 
A curriculum approach was presented in~\cite{soviany2021curriculum}, by opting for Cycle-GAN for the styling task, which enables the detector to adapt towards the target domain in a progressive way from easy to hard samples.
The pipeline developed in \cite{yu2022sc} manipulates the receptive field of Cycle-GAN to pay more attention to low-level style cues such as object details and its immediate context.
Munir et al.~\cite{munir2020thermal} also argue that low-level features work well when styling source images from the visible spectrum to target images from the infrared spectrum. 
A bidirectional styling approach was proposed in \cite{zhou2023ssda} by incorporating source-like target images and target-like source images into a student-teacher learning framework. 
In this regard, style transfer is a straightforward technique to alleviate domain discrepancy.
However, it remains limited in scenarios where a distribution shift occurs within the source and/or the target domains themselves, in which case style transfer may imitate cross-domain style partially.

\textbf{Feature alignment} approaches go beyond image-level and reason at feature-level instead. Adversarial training is the most frequent approach. So far, there is no consensus whether global or local feature alignment works best. 
In \cite{zhu2019adapting}, regions of interest (e.g., region proposals derived from a Faster-RCNN) are mined and their respective feature representations are subsequently aligned in an adversarial manner. 
Motivated by the fact that image-level feature alignment leads to noisy foreground-background alignment, and that instance-level alignment often involves background signals, the method in \cite{hsu2020every} pays more attention to foreground pixels with higher objectness and centerness scores (these latter two are infered from objectness and centerness maps via a fully-convolutional module). 
Other methods combine both style transfer and feature alignment such as \cite{hsu2020progressive}, where source images are translated to target style using a CycleGAN, then the intermediate stylized domain image features are aligned with the target features via adversarial learning. In \cite{rezaeianaran2021seeking}, instance proposals (Faster R-CNN) are first aggregated based on visual similarity into class-agnostic feature clusters via hierarchical agglomerative clustering. 

\textbf{Pseudo-label self-training} is another approach that mines confident detections from the target images and uses them as supervision to adapt the detector.
In SC-UDA \cite{yu2022sc}, multiple detections from the target are obtained by implementing a stochastic dropout into the detector's feature extraction layers during both training and testing phases. The detections with higher IoUs are then fused and leveraged to self-adapt the detector.
Confident pseudo-labels combined with hard augmentations are used to self-train the detector in \cite{roychowdhury2019automatic}. ConfMix~\cite{mattolin2023confmix} selects the most confident region in the target image and pastes it into the source image, which is redundant because the detector is already supervised with consistency loss on the source data during adaptation. 
DACA~\cite{mekhalfi2023daca} tackles this downside. Instead of mixing the confident target region with a source image, DACA augments it randomly (along with the pseudo-labels therein) to produce challenging regions and then joins these latter into a final image that can be used to adapt the detector. Therefore, DACA does not mix cross-domain images but instead composes informative target images to adapt the detector.

\subsection{Source-free unsupervised domain adaptation for object detection}

SF-UDA has attracted far less attention w.r.t UDA.
Most existing works in this setting opt for a teacher-student learning paradigm (see the second part of Tab.~\ref{tab:sota}).
For instance, in \cite{xiong2021source} an auxiliary target domain is created by perturbing target images.
Then the target and the auxiliary target images are fed to the teacher and student models (both pretrained on the source domain).
The student model admits region proposals (i.e., Faster R‐CNN) from the teacher model, and is trained with three consistency losses, namely image-level, instance-level, and category-level.
LODS~\cite{li2022source} adopts a similar concept by generating an auxiliary target domain via target image style enhancement.
This is achieved through a non-linear combination of the input target image and any other target image (i.e., fully connected layers and a ReLU applied on the image features derived by a pretrained VGG-16 model).
ASFOD~\cite{chu2023adversarial} argues that the source-pretrained detector tends to infer more predictions on hard target images that exhibit similarity to the source domain, while ignoring hard target images that are dissimilar to the source, yet they manifest high uncertainty, and score high variance during adaptation. Intuitively, the target domain is divided accordingly into source-similar and source-dissimilar subsets. Next, a student-teacher flow is adopted to align these two subsets in an adversarial learning. Finally, a mean-teacher step is achieved to finetune the detector.
LPU~\cite{chen2023exploiting} imposes two confidence thresholds on the pseudo-labels generated by the teacher model, namely a high threshold and a low threshold.
For the detections exceeding the high threshold, traditional mean-teacher framework is invoked.
As per the pseudo-labels exceeding the low threshold and below the high threshold, proposal soft training is carried out by feeding the (Faster-RCNN) proposals to the teacher model to obtain class predictions, which are then leveraged to self-train the student model with its own proposals.
Moreover, an IoU-mixup is conducted on the proposals in the spatial location neighborhood, to make the model aware of the relationship between neighboring proposals via contrastive learning.
SFOD-Mosaic~\cite{li2021free} employs self-training with pseudo-labels to adapt towards the target.
It assumes that the noise degree of the pseudo-detections is positively correlated with the mean self-entropy, indicating a potential to measure the predictions uncertainty.
Thus, positive and negative target samples are distinguished by searching for a confidence threshold from the higher to the lower scores, and self-entropy descent is observed when the mean self-entropy descends and hits the first local minimum, and adapts the detector accordingly.

\begin{table*}
\centering
\renewcommand{\arraystretch}{0.8}

\caption{
Overview of existing UDA and SF-UDA approaches.
The following notations are used in the table, namely Source: source dataset, Target: target dataset, C: Cityscapes, F: FoggyCityscapes, K: KITTI, S: Sim10K, P: Pascal VOC, W: Watercolor2k, B: BDD100k, Cl: Clipart. 
The proposed SF-DACA is the only data augmentation-driven approach in the SF-UDA setting so far.
}
\label{tab:sota}

\vspace{-2mm}
\resizebox{\linewidth}{!}{%
\begin{tabular}{lccccc}
\toprule
Method & Setting & In a nutshell & Base detector & Source $\to$ Target & Code \\
\toprule

%
%
ConfMix \cite{mattolin2023confmix} 
& UDA
& \makecell{Self-training \\ Cross-domain image mixing}  
& YOLOv5
& \makecell{C $\to$ F,  S $\to$ C \\ K $\to$ C}  
& \href{https://github.com/giuliomattolin/ConfMix}{\underline{link}} \\
\midrule

DACA \cite{mekhalfi2023daca} 
& UDA
& \makecell{Self-training \\ Target-only region mixing \\ Data augmentation}  
& YOLOv5
& \makecell{C $\to$ F \\ S $\to$ C \\ K $\to$ C}  
& \href{https://github.com/MohamedTEV/DACA}{\underline{link}} \\
\midrule

DA-Detect \cite{li2023domain} 
& UDA
& \makecell{Image-level and object-level adaptation \\ Adversarial mining for hard examples \\ Auxiliary domain generation} 
& Faster R-CNN 
& \makecell{C $\to$ F \\  C $\to$ K}  
& \href{https://github.com/jinlong17/DA-Detect}{\underline{link}} \\
\midrule

SC-UDA \cite{yu2022sc} 
& UDA
& \makecell{Intermediate domain generation with optimized CycleGAN \\ Iterative self-training \\ Uncertainty-based pseudo-label optimization} 
& Faster R-CNN 
& \makecell{C $\to$ F \\  S $\to$ C \\ K $\to$ C}  
& \ding{55} \\
\midrule

ViSGA \cite{rezaeianaran2021seeking} 
& UDA
& \makecell{Instance-level feature aggregation \\ Similarity-based RPN proposal clustering \\ Alignment via adversarial training} 
& Faster R-CNN 
& \makecell{C $\to$ F \\  S $\to$ C \\ K $\to$ C}  
& \ding{55} \\
\midrule


SOAP \cite{xiong2021source} 
& SF-UDA
& \makecell{Mean teacher \\ Perturbed target images}  
& Faster R-CNN 
& \makecell{C $\to$ F,  S $\to$ C \\ K $\to$ C, P $\to$ W}  
& \href{https://github.com/xiong233/SOAP}{\underline{link}} \\ 
\midrule

SFOD-Mosaic \cite{li2021free} 
& SF-UDA
& \makecell{Mining pseudo-label  threshold via self-entropy descent} 
& Faster R-CNN 
& \makecell{C $\to$ F,  S $\to$ C \\ K $\to$ C, C $\to$ B} 
& \ding{55} \\ 
\midrule

LODS \cite{li2022source} 
& SF-UDA
& \makecell{Mean teacher \\ Non-linear target style enhancement}
& Faster R-CNN 
& \makecell{C $\to$ F,  K $\to$ C \\ P $\to$ Cl, P $\to$ W} 
& \href{https://github.com/Flashkong/Source-Free-Object-Detection-by-Learning-to-Overlook-Domain-Style}{\underline{link}} \\ 
\midrule

ASFOD \cite{chu2023adversarial} 
& SF-UDA
& \makecell{Student-teacher \\ Adversarial learning \\ Source-similarity-based target  division}
& Faster R-CNN 
& \makecell{C $\to$ F,  S $\to$ C \\ K $\to$ C, C $\to$ B} 
& \href{https://github.com/ChuQiaosong/AASFOD}{\underline{link}} \\ 
\midrule

LPU \cite{chen2023exploiting} 
& SF-UDA
& \makecell{Mean teacher \\ Confident pseudo-labels \\ Proposal soft training on low-confident pseudo-detections}
& Faster R-CNN 
& \makecell{C $\to$ F,  S $\to$ C \\ K $\to$ C, C $\to$ B} 
& \ding{55} \\ 
\midrule

SF-UT \cite{hao2024simplifying}
& SF-UDA
& \makecell{Mean teacher \\ Adaptive Batch Normalization}
& Faster R-CNN 
& \makecell{C $\to$ F,  S $\to$ C \\ K $\to$ C} 
& \href{https://github.com/EPFL-IMOS/simple-SFOD}{\underline{link}} \\
\midrule

SF-DACA (ours)
& SF-UDA
& \makecell{Mean teacher \\ Target-only region mixing \\ Data augmentation}  
& Faster R-CNN 
& \makecell{C $\to$ F \\  S $\to$ C \\ K $\to$ C} 
& Soon \\
\bottomrule
\end{tabular}
}

\end{table*}

\section{Our method}\label{method}

\subsection{Overview}
Suppose a labeled source domain $D_s = \{({x}_s^i, {y}_s^i)\}_{i=1}^{N_s}$ and an unlabeled target domain $D_t = \{x_t^i\}_{i=1}^{N_t}$ that obey disjoint distributions, where ${x}_s^i \in \mathbb{R}^{H\times W \times3}$ and ${x}_t^i \in \mathbb{R}^{H\times W \times3}$ are RGB images randomly sampled from the source and target domains respectively, and ${y}_s^i = (b_s^i \in \mathbb{R}^{M_s \times 4}, c_s^i \in \mathbb{N}^{M_s  \times 1})_{i=1}^{N_s}$ denotes the groundtruth bounding box coordinates and categories of the objects included in the image $x_s$, where $M_s$ is the number of objects, and $C$ is the number of object classes.
An object detection model can be defined as a parametric function $\Phi_\Theta \colon \mathbb{R}^{H \times W \times 3} \to \mathbb{R}^{M \times 4} \times [1, \dots, C]^{M}$ with learnable parameters $\Theta$ that admits an RGB image $x$ as input and infers a list of $M$ detections.

The aim in classical UDA is to adapt an object detector that was pretrained on the source domain $D_s$ to the target domain $D_t$, while having access to the source data in $D_s$ during adaptation, which helps prevent catastrophic forgetting. In SF-UDA, the source data is unavailable, and the source-pretrained detector is adapted with target images $D_t = \{x_t^i\}_{i=1}^{N_t}$ only.

In order to cope with potential model collapse due to the lack of source data, in this paper we develop a student-teacher learning approach that assumes two source-pretrained models that have identical architectures, namely a teacher detector parameterized with $\Theta_{\textrm{tea}}$ and a student model parameterized with $\Theta_{\textrm{stu}}$. The role of $\Theta_{\textrm{stu}}$ is to adapt towards $D_t$, while the role of $\Theta_{\textrm{tea}}$ is to prevent $\Theta_{\textrm{stu}}$ from collapsing.

\subsection{Self-training with regional pseudo-labels}
The proposed SF-DACA takes advantage of self-training to approach the target domain. This indicates the mining of pseudo-labels from the target images and utilizing them as groundtruth to finetune the detector on the target. Yet, we argue that these pseudolabels should be (i) reliable, (ii) convey spatial context, and (iii) informative/challenging. The reliability implies that the detector is adapted with pseudo-labels that correspond to actual objects of interest, and the spatial context enables the detector to perceive the visual aspect of objects among other objects/background, while the informative/challenging property prevents the detector from overfitting the target images.       

To tackle all these three key properties, we propose to adapt the student model $\Phi_{\Theta_{\textrm{stu}}}$ with composite target images that are formed by leveraging regional pseudo-labels.
To this end, given a randomly sampled target image $x_t$, we feed it to the student model $\Theta_{\textrm{stu}}$ to infer a set of $M_{stu}$ detections ${y}_{stu} = (b_{stu} \in \mathbb{R}^{M_{stu} \times 4}, c_{stu} \in \mathbb{N}^{M_{stu} \times 1}, p_{stu} \in \mathbb{R}^{M_{stu} \times 1})$, where $p_{stu}$ is the detection confidence. 
First, to make use of reliable pseudo-labels with visual context, we divide the target image $x_t$ into four spatial regions $r_{tl}, r_{tr}, r_{bl}, r_{br}$, which respectively stand for the top left, top right, bottom left, and bottom right image regions that constitute the whole target image $x_t = r_{tl} \cup r_{tr} \cup r_{bl} \cup r_{br}$.

Next, we select the most confident image region based on the average detection confidence in that image portion $p_r = \frac{1}{n_r} \sum_{i=1}^{n_r} p_i$ where $n_r$ is the number of regional detections and $p_i$ are their respective confidences. In particular, in each image region, we calculate the average detection confidence of all the detections whose bounding box centers fall into it, and we retain only the most confident region $r_{conf}$ with the highest average detection confidence, as in Eq.~\ref{eq_bestregion}, along with its pseudo-labels,   

\vspace{-13pt}
\begin{equation}
r_{conf} = \{r : r\in (r_{tl}, r_{tr}, r_{bl}, r_{br}) \text{ if } p_r = \max (P_{stu}) \},
\label{eq_bestregion}
\end{equation}

\noindent where $P_{stu} = (\bar p_{tl}, \bar p_{tr}, \bar p_{bl}, \bar p_{br})$ stands for the region-wise average detection confidences. 

Second, to create challenging samples for the detector to learn from, the confident image region $r_{conf}$ and its respective detections are augmented randomly four times (see Fig.~\ref{fig:pipeline}), and the augmented versions of that region are joined into a single composite image,  

\vspace{-13pt}
\begin{equation}
x_{comp}, b_{comp}, c_{comp}  = \cup_{i=1}^{4}S_i(r_{conf}, b_{conf}, c_{conf}), 
\label{eq_composite}
\end{equation}
\vspace{-13pt}

\noindent where $S_i$ is the random augmentations function, $x_{comp}$ is the composed challenging target image, $b_{comp}$ is the set of augmented bounding boxes and $c_{comp}$ is the set of classes, $b_{conf}$ and $c_{conf}$ are the set of bounding boxes of the confident target region and their object classes respectively.

Afterwards, the composite image $x_{comp}$ is fed to the student model for inference, and the consistency loss $\ell_s$ between the inferred detections $y_{inf} = (b_{inf} \in \mathbb{R}^{M_{inf} \times 4}, c_{inf} \in \mathbb{N}^{M_{inf} \times 1})$ and the composite detections $y_{comp} = (b_{comp} \in \mathbb{R}^{M_{comp} \times 4}, c_{comp} \in \mathbb{N}^{M_{comp} \times 1})$ is calculated,     

\vspace{-13pt}
\begin{equation}
\ell_s = \mathcal{L}(y_{comp}, y_{inf}),
\label{eq_ls}
\end{equation}
\vspace{-13pt}

\noindent where $ \mathcal{L}(.)$ is the loss function of the adopted detection model.

\subsection{Adaptation}
The adaptation process of the student model $\Theta_{stu}$ follows a selective augmentation approach of image regions and their pseudo-detections. Ideally, these regions should comprise only true positives. However, this does not hold as image regions entail outlier false positives as well. The augmentation step further accumulates the contribution of these outliers, leading the detector to collapse eventually. Moreover, the knowledge of the student model gained on the source domain in the pretraining phase is prone to be forgotten. 

\begin{figure*}

\centerline{\includegraphics[width=0.9\textwidth]{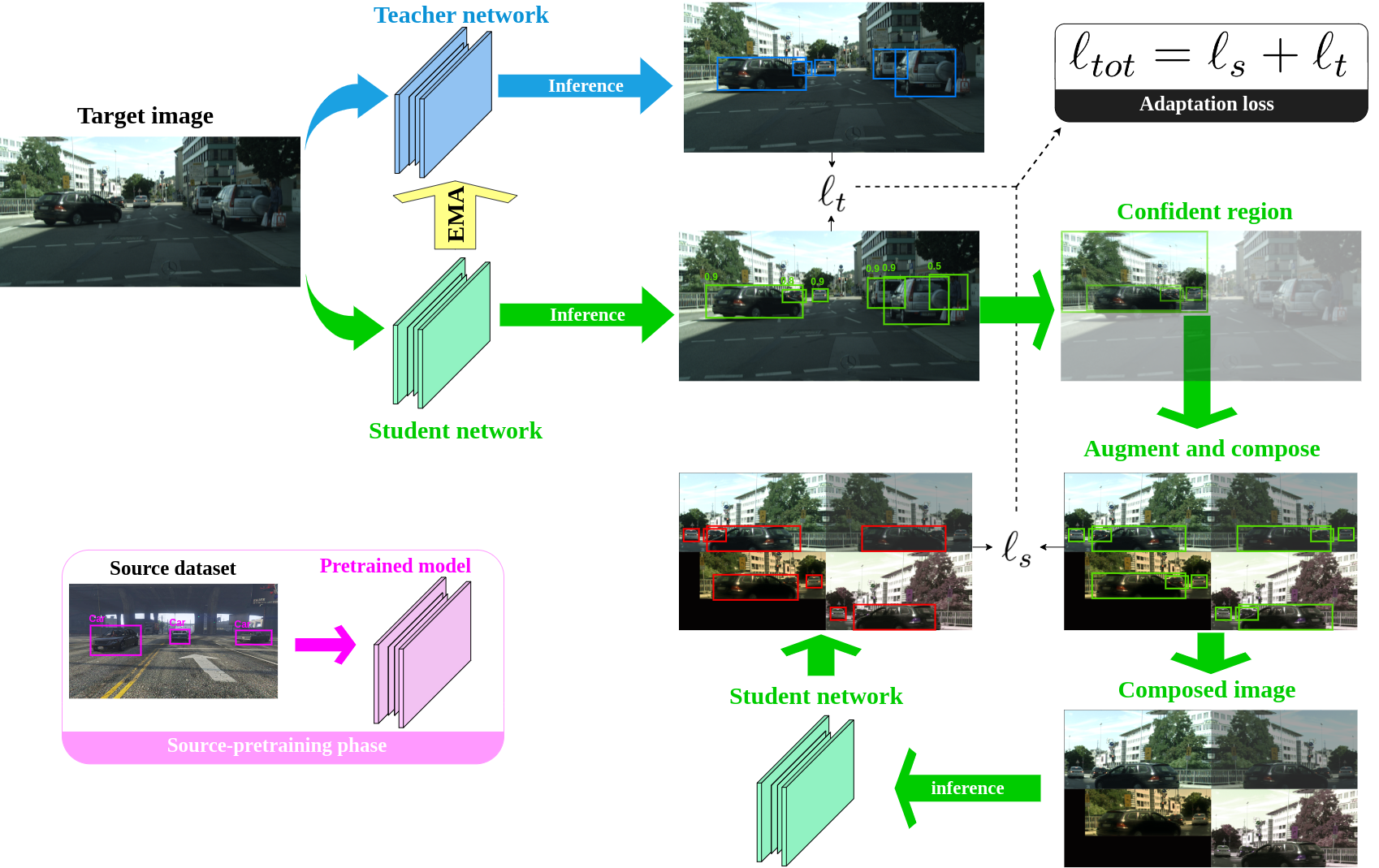}}

\caption{Proposed SF-DACA. The student and the teacher models posses the same source-pretrained architecture (bottom-left). The green flow indicates adaptation towards the target via self-training with augmented and composed confident target regions. The blue flow refers to the teacher supervision to refrain the student from collapsing. Yellow flow ensures that the teacher model distills target knowledge from the student progressively along the adaptation progress.}

\label{fig:pipeline}
\end{figure*}

To cope with this bottleneck, we back up the adaptation process with another teacher model $\Theta_{tea}$ also pretrained on the source data. Although source data is inaccessible, the source-pretrained model does convey rich information about the source domain. As depicted in Fig.~\ref{fig:pipeline}, the target image $x_t$ is fed to both the student and teacher models. The teacher receives the whole target image and delivers $M_{tea}$ pseudo-labels  ${y}_{tea} = (b_{tea} \in \mathbb{R}^{M_{tea} \times 4}, c_{tea} \in \mathbb{N}^{M_{tea} \times 1}, p_{tea} \in \mathbb{R}^{M_{tea} \times 1})$, where $p_{tea}$ is the detection confidence. Simultaneously, as described previously, the student receives the whole target image and discerns $M_{stu}$ pseudo-labels  ${y}_{stu} = (b_{stu} \in \mathbb{R}^{M_{stu} \times 4}, c_{stu} \in \mathbb{N}^{M_{stu} \times 1}, p_{stu} \in \mathbb{R}^{M_{stu} \times 1})$. The consistency loss can then be calculated as in Eq.~\ref{eq_lt}. Finally, the total adaptation loss $\ell_{tot}$ is given in Eq.~\ref{eq_ltot}. 

\vspace{-13pt}
\begin{equation}
\ell_t = \mathcal{L}(y_{tea}, y_{stu}),
\label{eq_lt}
\end{equation}
\vspace{-13pt}

\vspace{-13pt}
\begin{equation}
\ell_{tot} = \ell_s + \ell_t.
\label{eq_ltot}
\end{equation}
\vspace{-13pt}

The student model is supervised with $\ell_s$ to approach the target domain, and distills knowledge from the teacher model via $\ell_t$ to prevent catastrophic forgetting. To controllably progress the knowledge of the teacher model towards the target, we adopt an exponential moving average (EMA) \cite{ema} to pass knowledge from the student to the teacher by partially sharing its parameters,

\vspace{-13pt}
\begin{equation}
\theta_{tea} = \alpha \cdot \theta_{tea} + (1 - \alpha) \cdot \theta_{stu},
\label{eq_lema}
\end{equation}
\vspace{-13pt}

\noindent where $\alpha$ is a hyperparameter that controls the amount of student parameters that are shared to the teacher. This step is invoked at the end of each adaptation epoch as detailed in Pseudocode.~\ref{pseudocode}.

\IncMargin{1.5em}
\begin{algorithm}[t]
	\SetKwData{Left}{left}
	\SetKwData{This}{this}
	\SetKwData{Up}{up}
	\SetKwFunction{Union}{Union}
	\SetKwFunction{FindCompress}{FindCompress}
	\SetKwInOut{Input}{input}
	\SetKwInOut{Output}{output}  
\Input{
$\theta_{stu}$ and $\theta_{tea}$,
target images ${x}_t^i$,
random augmentations $S$,
loss function $\mathcal{L}(.)$,
epochs $N_{ep}$,
iterations $N_{it}$.
}

\Output{Adapted model $\theta_{stu}$.}
\BlankLine

\For{$ep = 1, \dots , N_{ep}$}
{

\For{$i = 1, \dots , N_{it}$}
{

Feed ${x}_t^i$ to  $\theta_{stu}$ and obtain $y_{stu}$,

Divide the image equally into $2\text{x}2$ layout,

Select $r_{conf}$ (Eq.~\ref{eq_bestregion})),

Augment and compose (Eq.~\ref{eq_composite}),

Feed $x_{comp}$ to $\theta_{stu}$ and obtain $y_{inf}$,

Compute $\ell_s  \leftarrow \mathcal{L}(y_{comp}, y_{inf})$,

Feed ${x}_t^i$ to  $\theta_{tea}$ and obtain $y_{tea}$,

Compute $\ell_t  \leftarrow \mathcal{L}(y_{tea}, y_{stu})$,

Compute and minimize $\ell_{tot} = \ell_s + \ell_t$,

}

Update teacher with EMA: $\theta_{tea} \leftarrow \alpha \cdot \theta_{tea} + (1 - \alpha) \cdot \theta_{stu}$,

}
\textbf{Return}: Adapted student parameters $\theta_{stu}$.

\caption{Pseudocode of SF-DACA.}
\label{pseudocode}
\end{algorithm}
\DecMargin{1.5em}

\section{Experiments} \label{experiments}
\subsection{Datasets}

We validated SF-DACA on four traffic scenes datasets: Cityscapes~\cite{cityscapes}, FoggyCityscapes~\cite{foggycityscapes}, Sim10K~\cite{sim10k}, and KITTI~\cite{kitti}.
Cityscapes (C) comprises real-world urban scenes captured across different cities. It includes object bounding box annotations for eight categories: Person, Rider, Car, Truck, Bus, Train, Motorcycle, and Bicycle. 
FoggyCityscapes (F) is a modified version of Cityscapes, generated by applying a synthetic fog filter to the original Cityscapes images. 
Sim10K (S) consists of synthetic images generated with the GTA-V game engine.
KITTI (K) is a real-world traffic scenes dataset.
To be consistent with previous SF-UDA works~\cite{xiong2021source, li2021free, li2022source, chu2023adversarial, chen2023exploiting, hao2024simplifying}, we considered three adaptation scenarios.
The first one, denoted as C$\to$F, considers Cityscapes as the source domain and FoggyCityscapes as the target domain, addressing the adaptation to different weather conditions (clear vs foggy). It encompasses eight object categories.
The second one, K$\to$C, involves cross-camera adaptation for the Car category, with KITTI as the source and Cityscapes as the target.
The third scenario, S$\to$C, focuses on synthetic-to-real adaptation for the Car category, using Sim10K as the source domain and Cityscapes is the target. 

\subsection{Implementation details}
For the sake of fairness with existing SF-UDA approaches, we consider Faster R-CNN \cite{faster} with a VGG16 backbone pretrained on ImageNet \cite{simonyan2014very} as the base detector.
The images are resized such that the shorter size is set to 600 pixels.

As optimizer, we opt for stochastic gradient descent with a learning rate of 1e-3 and a momentum of 0.9.
In the pretraining phase, the model is trained for 5 epochs on the Cityscapes dataset when dealing with the C$\to$F scenario, for 1 epoch on KITTI and Sim10K regarding K$\to$C and S$\to$C, respectively.
As per the adaptation phase, the detector is trained for 12 epochs for all three scenarios.  
The layout of 2$\times$2 is considered to build the composite target image according to our findings in our previous work \cite{mekhalfi2023daca}. This entails that the most confident region of the target image is augmented 4 times randomly, and the 4 augmented versions are put together into a 2$\times$2 layout to form an image as shown at the bottom right of Fig.~\ref{fig:pipeline}. An ablation on the most effective combination of data augmentations can be found in~\cite{mekhalfi2023daca}.
The mean average precision (mAP) and the average precision (AP) with an IoU threshold of 0.5 are reported as evaluation metrics.

\subsection{Results and discussion}

Tables \ref{tab:c2f} and \ref{tab:k2c_s2c} present quantitative results comparing \daca~with several source-free competitors based on the same detector and backbone, to ensure fairness.
Specifically, we compare against SF-UT \cite{hao2024simplifying}, LPU \cite{chen2023exploiting}, ASFOD \cite{chu2023adversarial}, LODS \cite{li2022source}, SFOD \cite{li2021free}, and SOAP \cite{xiong2021source}. 
The Source and Oracle notations (rows 1 and 2 in the tables) denote the cases when the detector is trained only on the source data, and only on the target data, indicating the lower bound when the model is not adapted and the upper bound based on target groundtruth. 

The results of the C$\to$F adaptation scenario are summarized in Tab. \ref{tab:c2f}.
Compared to the Source baselines, our method achieves a 9.50\% mAP gain (row 10 in the table). Bicycle is the category that achieves the most gain w.r.t the source-only model, and Bus yields the lowest improvement. On the other hand, Bicycle and Rider classes are only 1.30\% and 1.50\% mAP away from the Oracle performance, and Bus class remains the least adaptable class among all (-19.00\% mAP loss w.r.t the Oracle), followed by the Train class (-13.90\%). 
Interestingly, the Train category reports a very low score (1.70\%) based on the source-only model. This implies that the adaptation starts with a model that possesses marginal knowledge about this class, posing self-adaptation issues. Note that this is the most difficult adaptation scenario due to the foggy weather and the multiple classes encountered in urban scenes. Compared with prior works, our method ranks third respectively after SF-UT \cite{hao2024simplifying} and LPU \cite{chen2023exploiting} mainly due to the low scores observed for the Bus and Train classes. By contrast, on the Car, Rider and Person categories, our method ranks first with improvements of 3.00\%, 3.20\% and 1.00\% compared with SF-UT \cite{hao2024simplifying}.

Concerning the K$\to$C, and S$\to$C adaptation cases, the scores are given in Tab. \ref{tab:k2c_s2c}. Remarkable improvements (+18.20\% and +22.90\%) are incurred for K$\to$C and S$\to$C, respectively. Compared to the state-of-the-art, for the K$\to$C our method ranks first among all with a large margin, followed by LPU \cite{chen2023exploiting} then SF-UT \cite{hao2024simplifying}. As per S$\to$C, our method ranks first, followed by SF-UT \cite{hao2024simplifying} then LPU \cite{chen2023exploiting}. Although the synthetic-to-real (S$\to$C) scenario starts from a lower score (36.70\%) compared to the cross-camera case (K$\to$C, with 38.60\%), its score post adaptation is higher (59.60\% vs 56.80\%), suggesting that domain discrepancy is larger in the cross-camera adaptation. Note that the adaptation gain in these two benchmarks remains much higher than the Car class of the C$\to$F (18.20\% and 22.90\% vs 10.60\%) mainly due to the severe domain shift in this latter case.

\begin{figure*}[htbp]
\centering
    \begin{tabular}{c c}
        Before SF-DACA (source-only) & 
        After SF-DACA \\

        \raisebox{4.5\normalbaselineskip}[0pt][0pt]{\rotatebox[origin=c]{90}{\footnotesize Cityscapes $\to$ FoggyCityscapes}}
        \includegraphics[width=7.3cm]{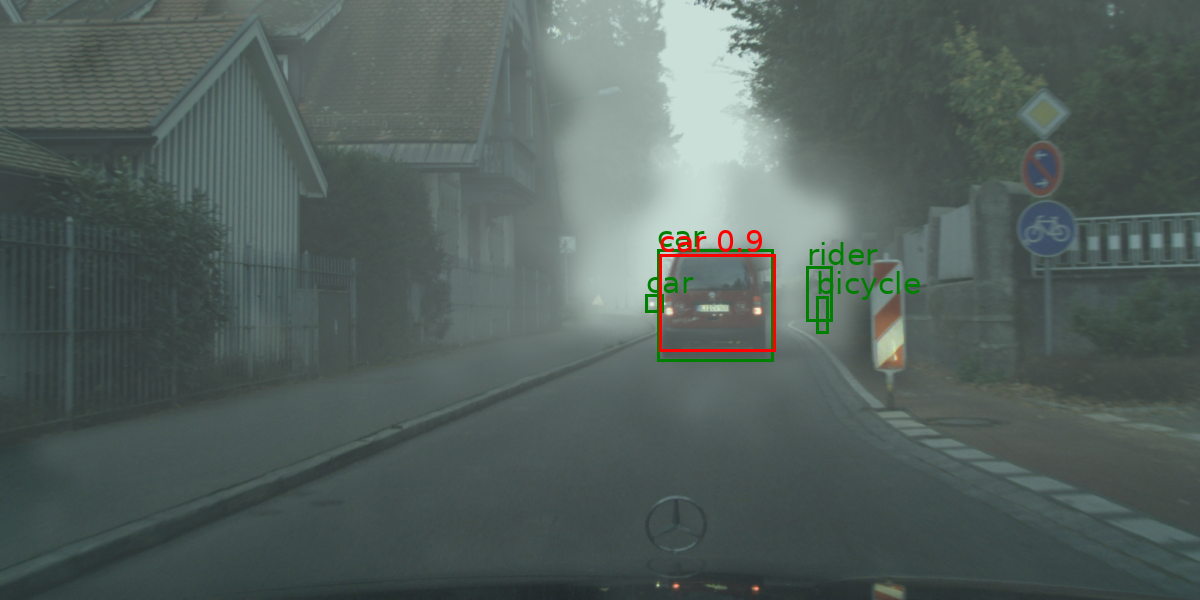} &
        \includegraphics[width=7.3cm]{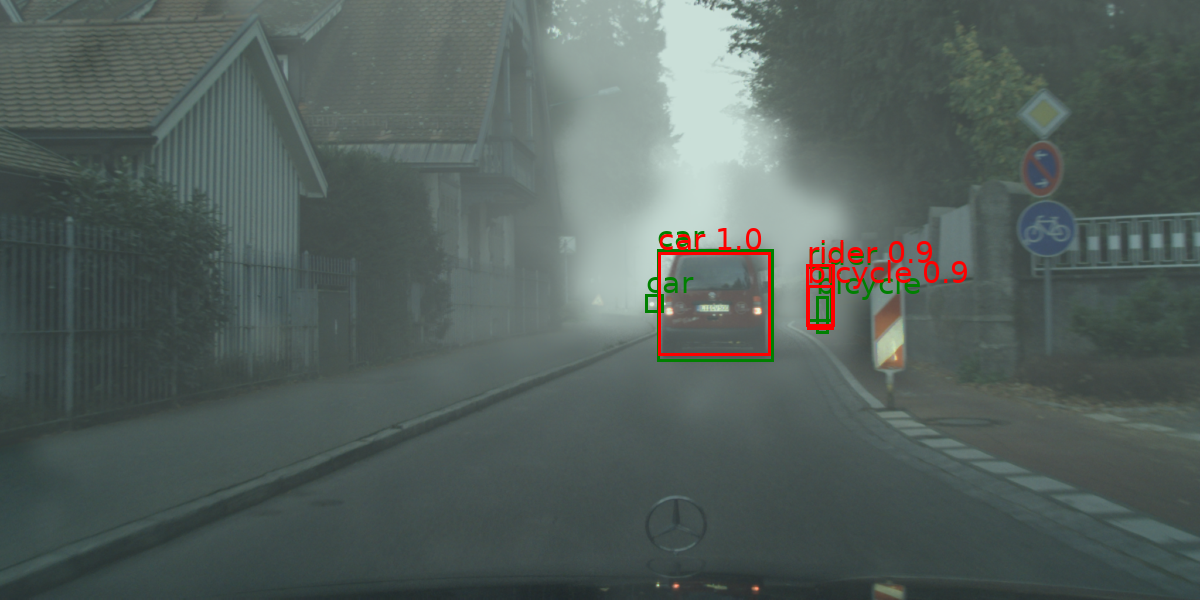} \\
        \raisebox{4.5\normalbaselineskip}[0pt][0pt]{\rotatebox[origin=c]{90}{\footnotesize Cityscapes $\to$ Foggycityscapes}}
        \includegraphics[width=7.3cm]{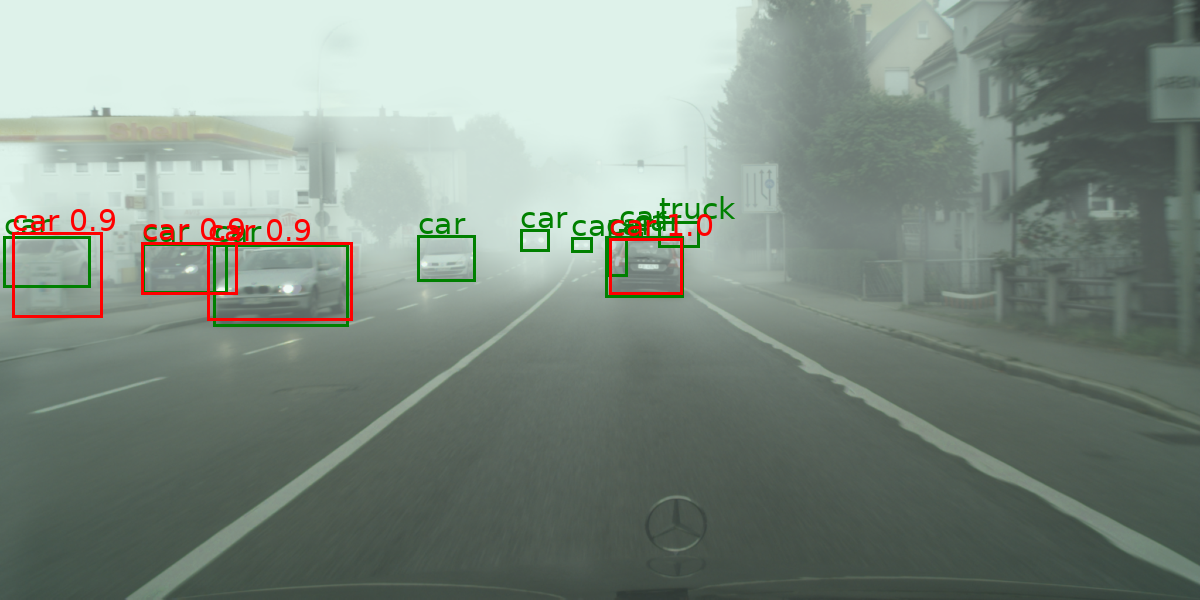}  &
        \includegraphics[width=7.3cm]{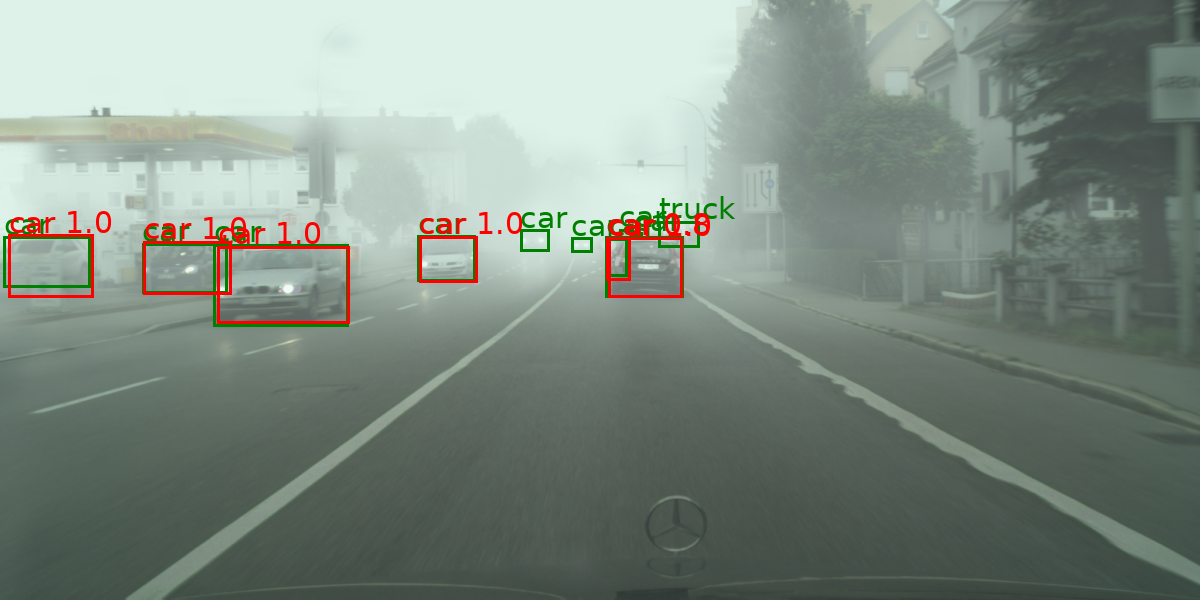} \\

        \raisebox{4.5\normalbaselineskip}[0pt][0pt]{\rotatebox[origin=c]{90}{\footnotesize KITTI $\to$ Cityscapes}}
        \includegraphics[width=7.3cm]{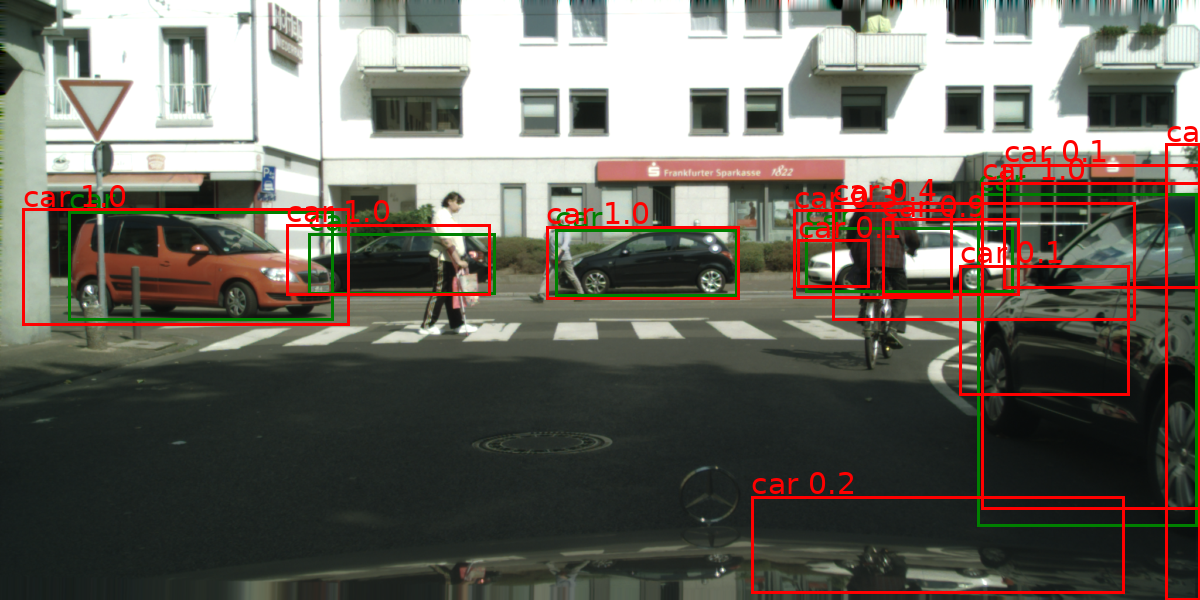}  &
        \includegraphics[width=7.3cm]{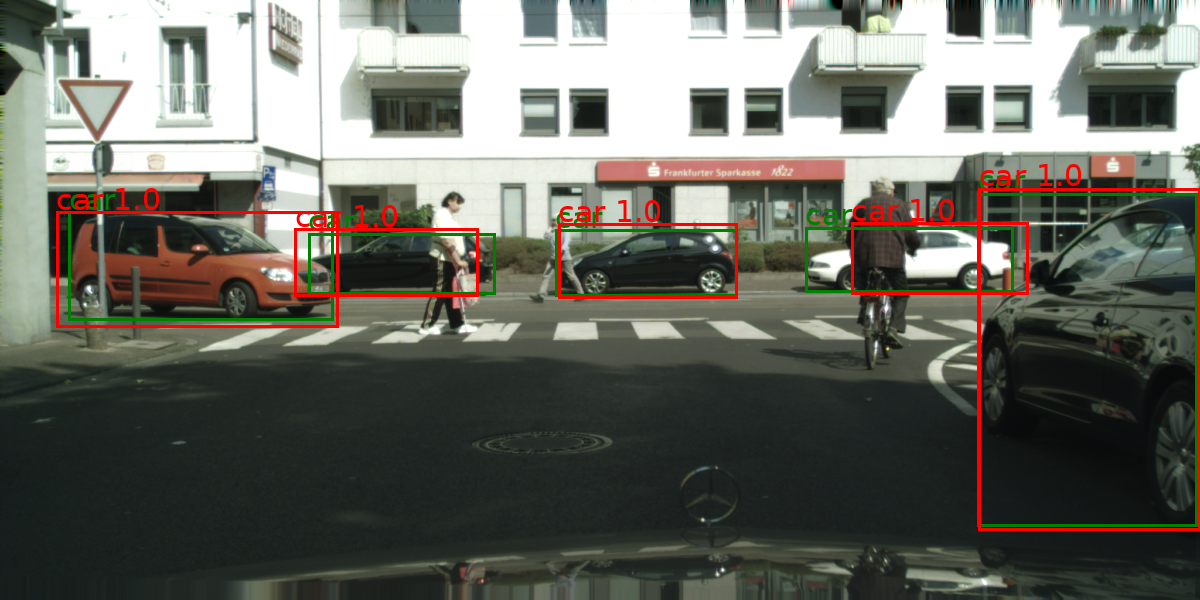} \\
        \raisebox{4.5\normalbaselineskip}[0pt][0pt]{\rotatebox[origin=c]{90}{\footnotesize KITTI $\to$ Cityscapes}}               
        \includegraphics[width=7.3cm]{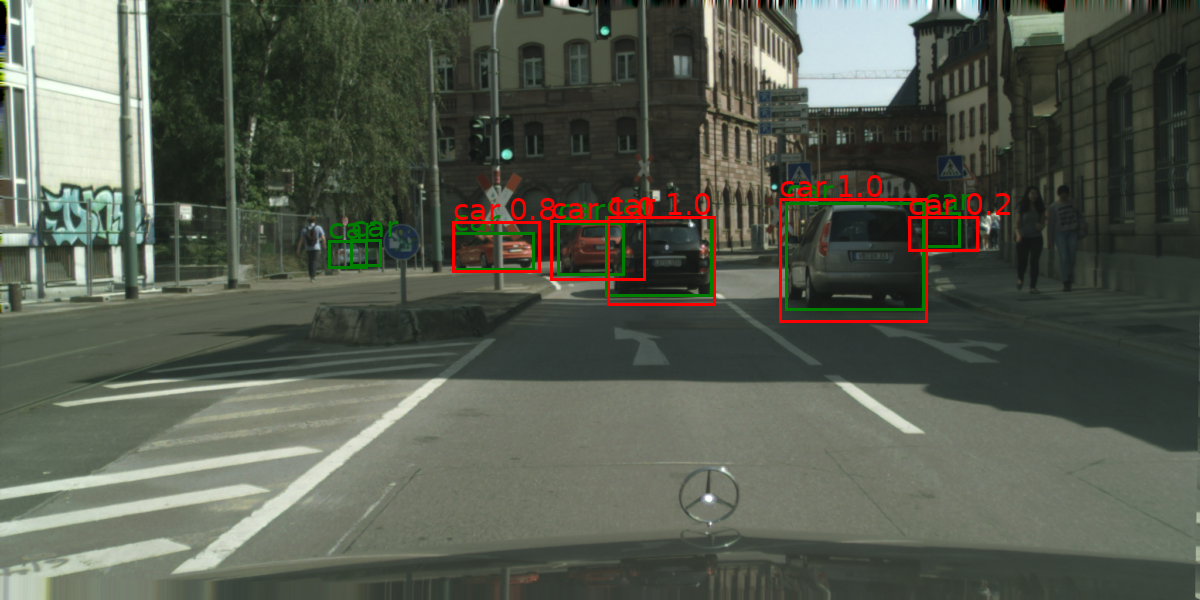}  &
        \includegraphics[width=7.3cm]{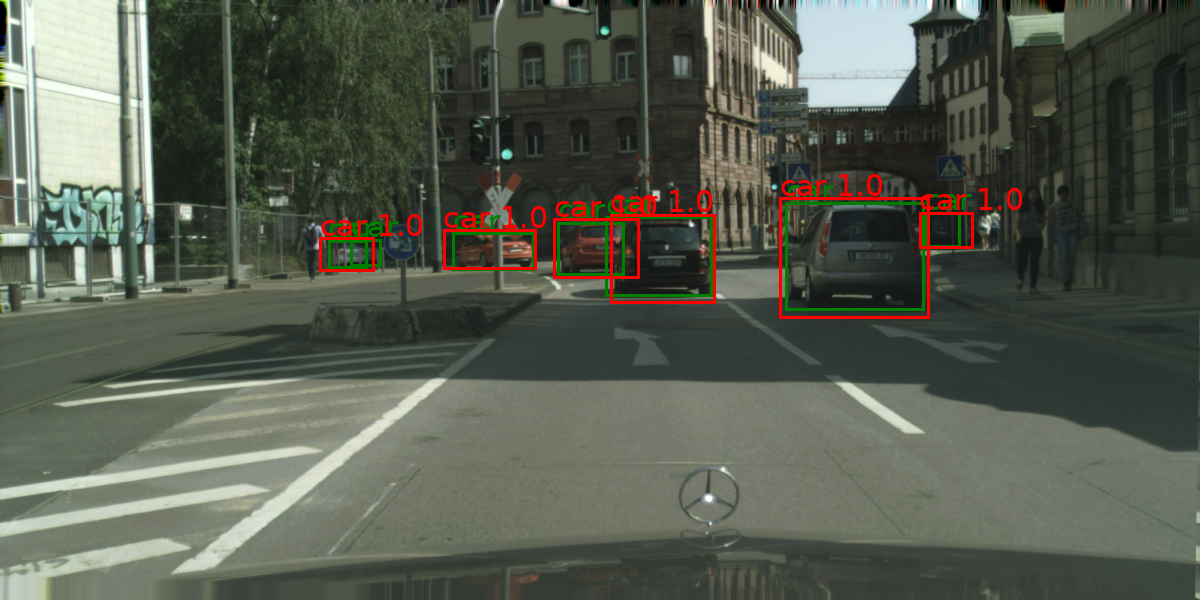} \\

        \raisebox{4.5\normalbaselineskip}[0pt][0pt]{\rotatebox[origin=c]{90}{\footnotesize SIM10K $\to$ Cityscapes}}
        {\includegraphics[width=7.3cm]{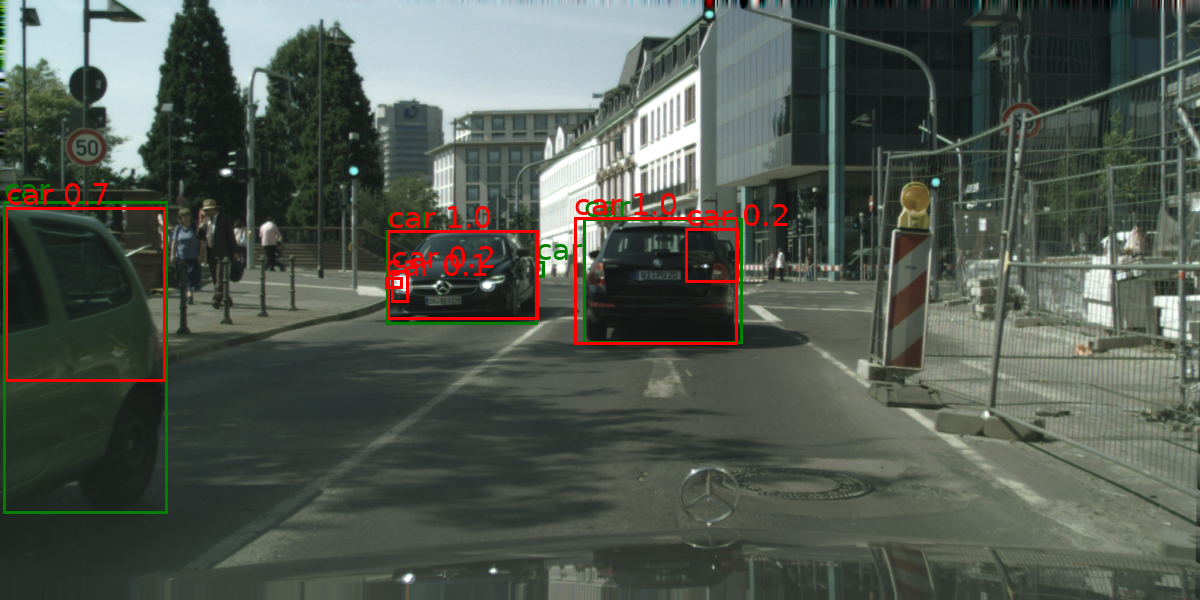}}  &
        {\includegraphics[width=7.3cm]{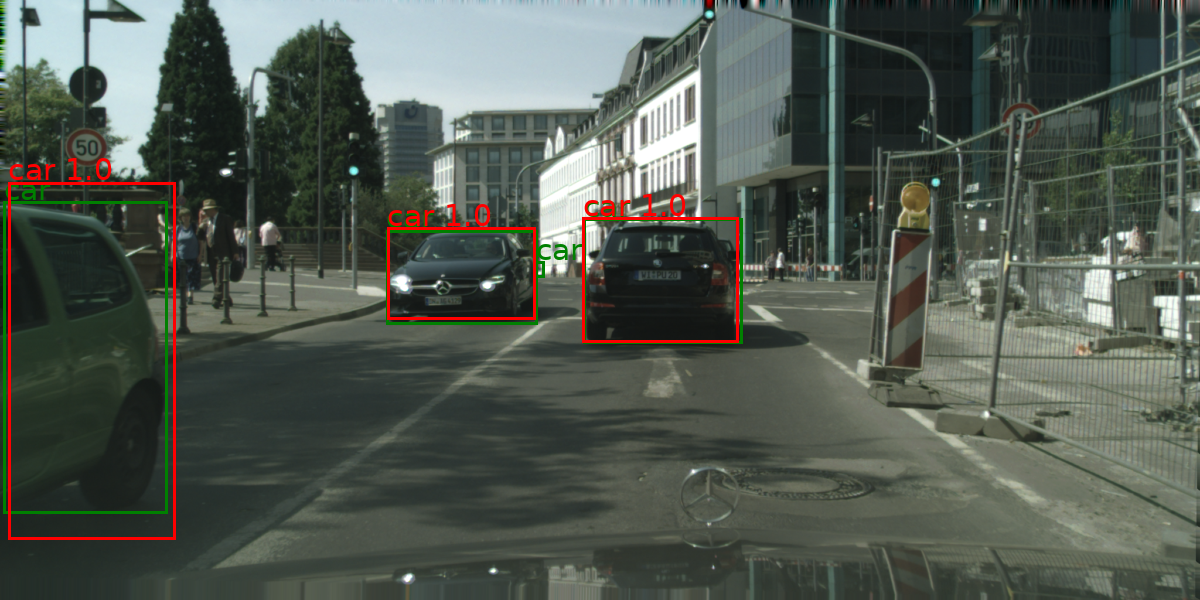}} \\
        \raisebox{4.5\normalbaselineskip}[0pt][0pt]{\rotatebox[origin=c]{90}{\footnotesize SIM10K $\to$ Cityscapes}}
        {\includegraphics[width=7.3cm]{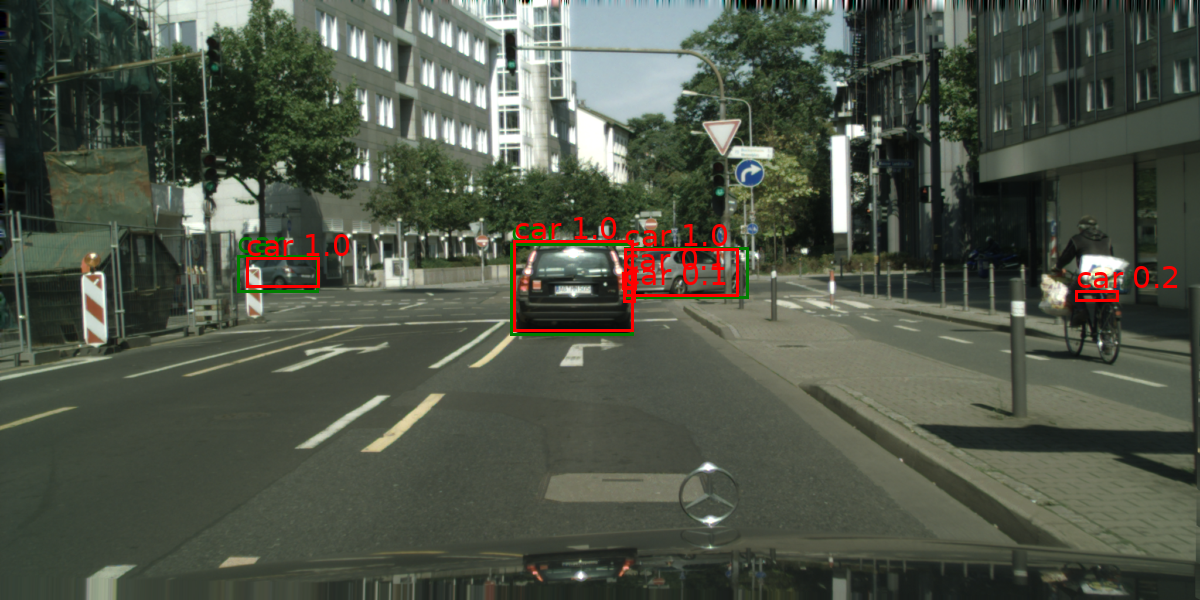}}  &
        {\includegraphics[width=7.3cm]{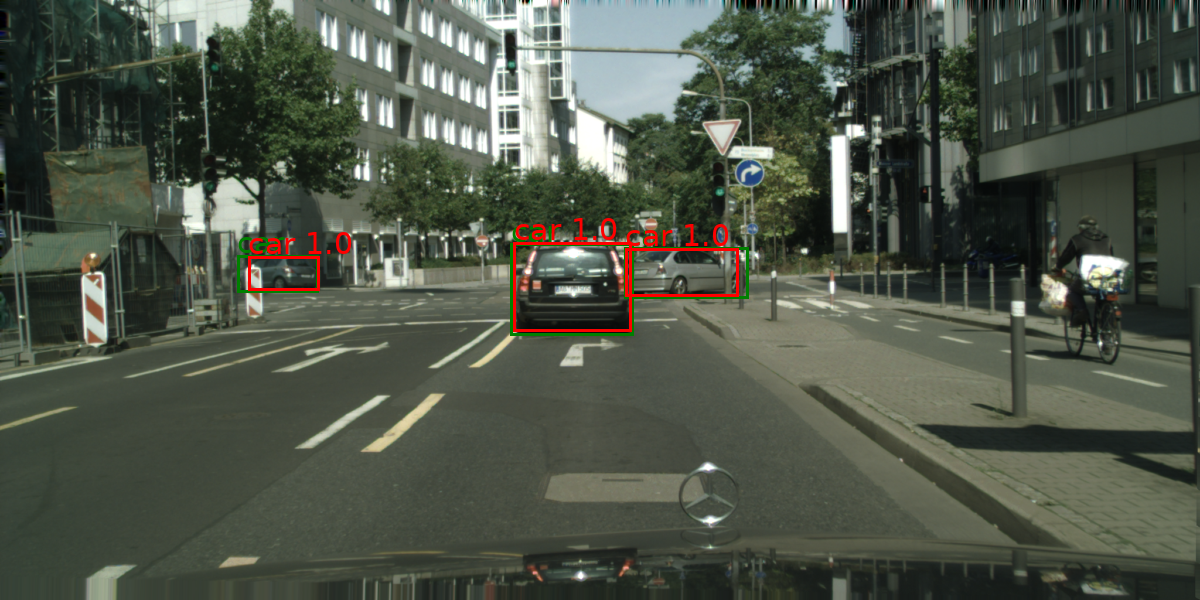}}
    \end{tabular}

\caption{
Qualitative examples from the C$\to$F (top 2 rows), K$\to$C (middle 2 rows) and S$\to$C (bottom 2 rows) adaptation scenarios. Detections of source-pretrained model (left) versus predictions of target-adapted model via SF-DACA (right). Green bounding boxes refer to the groundtruth, and red ones refer to the detections. Detected object class and confidence are provided on the top left of each instance. Best viewed in color.
}

\label{fig:qualitative_examples}
\end{figure*}

Qualitative examples before and after adaptation are illustrated in Fig.~\ref{fig:qualitative_examples}, for C$\to$F, K$\to$C and S$\to$C scenarios respectively, where less false positives, more true positives, and better fitted bounding boxes and higher confidences are observed post adaptation with \daca.

\begin{table*}
\renewcommand{\arraystretch}{0.8}
\tabcolsep 9pt
\centering

\caption{
Results on the Cityscapes to FoggyCityscapes adaptation scenario (C$\to$F). 
``Source only'' (row 1) refers to the lowerbound scores obtained by the source-pretrained model prior to adaptation.
``Oracle'' (row 2) indicates the upperbound scores by training the model on the target groundtruth.
Row 10 reports the gain w.r.t the Source-only (row 1).
Best performances are highlighted in bold.
}
\label{tab:c2f}

\vspace{-2mm}
\resizebox{\textwidth}{!}{%
\begin{tabular}{rlrrrrrrrrr}
\toprule
& Method & Person & Car & Train & Rider & Truck & Motorcycle & Bicycle & Bus & mAP \\
\toprule

\color{gray} \scriptsize 1 & Source-only & 35.20 & 51.30 & 1.70 & 41.90 & 12.90 & 18.70 & 29.40 & 32.20 & 27.90 \\

\color{gray} \scriptsize 2 & Oracle & 48.60 & 67.10 & 25.80 & 52.70 & 35.10 & 38.80 & 44.90 & 53.70 & 45.90 \\
\midrule

\color{gray} \scriptsize 3 & SOAP \cite{xiong2021source}  & 35.90 & 48.40 & 24.30 & 45.00 & 23.90 & 31.80 & 37.90 & 37.20 & 35.50\\ 

\color{gray} \scriptsize 4 & SFOD \cite{li2021free}  & 33.20 & 44.50 & 22.20 & 40.70 & 25.50 & 28.40 & 34.10 & 39.00 & 33.50\\

\color{gray} \scriptsize 5 & LODS \cite{li2022source} & 34.00 & 48.80 & 19.60 & 45.70 & 27.30 & 33.20 & 37.80 & 39.70  & 35.80\\ 

\color{gray} \scriptsize 6 & ASFOD \cite{chu2023adversarial} & 32.30 & 44.60 & 29.00 & 44.10 & 28.10 & 31.80 & 38.90 & 34.30  & 35.40\\

\color{gray} \scriptsize 7 & LPU \cite{chen2023exploiting} & 39.00 & 55.40 & 21.20 & 50.30 & 24.00 & 30.30 & \textbf{44.20} & 46.00  & 38.80\\

\color{gray} \scriptsize 8 & SF-UT \cite{hao2024simplifying} & 40.90 & 58.90 & \textbf{50.20} & 48.00 & \textbf{29.60} & \textbf{36.20} & 44.10 & \textbf{51.90} & \textbf{45.00}\\
\midrule

\color{gray} \scriptsize 9 & SF-DACA (ours) & \textbf{41.90} & \textbf{61.90} & 11.90 & \textbf{51.20} & 24.80 & 29.00 & 43.60 & 34.70 & 37.40 \\

\color{gray} \scriptsize 10 & Gain w.r.t. row 1 & \impp{6.70} & \impp{10.60} & \impp{10.20} & \impp{9.30} & \impp{11.90} & \impp{10.30} & \impp{14.20} & \impp{2.50} & \impp{9.50} \\


\bottomrule

\end{tabular}
}
\end{table*}

\vspace{-10pt} 

\begin{table}[t]
\renewcommand{\arraystretch}{0.8}
\caption{
Results on the KITTI to Cityscapes (K$\to$C) and the SIM10K to Cityscapes (S$\to$C) adaptation scenarios. 
Source (row 1) refers to the lowerbound scores obtained by the source-pretrained model prior to adaptation.
Oracle (row 2) indicates the upperbound scores by training the model on the target groundtruth.
Row 10 reports the gain w.r.t the Source-only (row 1).
Best performances are highlighted in bold.
}
\label{tab:k2c_s2c}
\centering
\tabcolsep 10pt
\resizebox{.9\columnwidth}{!}{%
\begin{tabular}{rlrr}
\toprule
& Method & K$\to$C & S$\to$C \\
\toprule
\color{gray} \scriptsize 1 & Source only & 38.60 & 36.70 \\
\color{gray} \scriptsize 2 & Oracle & 67.70 & 67.70 \\
\midrule
\color{gray} \scriptsize 3 & SOAP \cite{xiong2021source} & 41.90 & 40.80 \\
\color{gray} \scriptsize 4 & SFOD \cite{li2021free} & 44.60 & 43.10 \\
\color{gray} \scriptsize 5 & LODS  \cite{li2022source} & 43.90 & - \\
\color{gray} \scriptsize 6 & ASFOD \cite{chu2023adversarial} & 44.90 & 44.00 \\
\color{gray} \scriptsize 7 & LPU \cite{chen2023exploiting} & 48.40 & 47.30 \\
\color{gray} \scriptsize 8 & SF-UT \cite{hao2024simplifying} & 46.20 & 55.40 \\
\midrule
\color{gray} \scriptsize 9 & SF-DACA (ours) & \textbf{56.80} & \textbf{59.60} \\
\color{gray} \scriptsize 10 & Gain w.r.t. row 1 & \impp{18.20} & \impp{22.90} \\
\bottomrule
\end{tabular}
}
\end{table}

\subsection{Ablation study}

\noindent\textbf{Pseudo-label selection confidence.} 
Prior to selecting the most confident region with the highest confidence, all the detections inferred in the image undergo a selection step to filter our the likely false positives. This is done by simply imposing a selection confidence threshold.  
We report the adaptation scores for all the three adaptation cases in  Tab.~\ref{tab:pseudo_conf} according to different pseudo-label selection confidence values.
In particular, in the C$\to$F case, the mAP improves as the confidence increases up to a confidence threshold of 0.5, exhibiting a downward trend afterwards.
This suggests that pseudo-labels with low confidence below the value of 0.5 contain many false positives that misguide the adaptation.
Increasing the confidence beyond the value of 0.5 results in confident detections but at the cost of less true positives that are useful for the adaptation. Thus, we opt for a trade-off threshold of 0.5 as default choice.

Regarding the K$\to$C and the S$\to$C scenarios, the adaptation score manifests an upward trend along with the pseudo-label confidence. This is traced back to the fact these two benchmarks envision a one class adaptation problem, whereas in the earlier C$\to$F benchmark, the studied threshold is imposed on all the eight object classes. Therefore, for these two scenarios, the default threshold is set to 0.95.

\begin{table}[t]
\centering
\renewcommand{\arraystretch}{0.8}
\tabcolsep 6pt

\caption{Effect of the pseudo-label selection confidence.}
\label{tab:pseudo_conf}

\vspace{-2mm}
\resizebox{.8\columnwidth}{!}{%
\begin{tabular}{rcccc}
\toprule
& Selection conf. & C$\to$F & K$\to$C & S$\to$C \\
\toprule
\color{gray} \scriptsize 1 & 0.10 & 16.50 & 34.70 & 27.60 \\
\color{gray} \scriptsize 2 & 0.30 & 31.30 & 41.30 & 42.20 \\
\color{gray} \scriptsize 3 & 0.50 & \textbf{36.10} & 45.60 & 48.40 \\
\color{gray} \scriptsize 4 & 0.70 & 35.70 & 46.90 & 50.50 \\
\color{gray} \scriptsize 5 & 0.90 & 32.60 & 54.30 & 57.30 \\
\color{gray} \scriptsize 6 & 0.95 & 32.30 & \textbf{56.80} & \textbf{59.60} \\
\bottomrule

\end{tabular}
}
\end{table}

\noindent\textbf{Amount of shared student knowledge.} 
Note that the scores reported in the previous ablations were conducted with an EMA coefficient (Eq.~\ref{eq_lema}) of 0.9. Here, we fix the detection confidence according to the best values obtained earlier for each scenario and we further study the effect of $\alpha$ on the results as given in Tab.~\ref{tab:alpha} for all the three benchmarks. 
In all scenarios, the behaviors are about the same, where the performance increases as the $\alpha$ value is raised, up to a certain point ($\alpha$ of 0.95 in the C$\to$F benchmark and 0.9 otherwise) where it starts to decline afterwards. This suggests that sharing too much knowledge from the student to the teacher (i.e., small $\alpha$) leads the teacher to lose its own prior knowledge, causing the student to learn improperly from the teacher. By contrast, sharing a small amount of weights from the student to the teacher (i.e., large $\alpha$) causes the teacher to stagnate and holds the student from progressing its knowledge towards the target domain. Therefore, default values of 0.95 for the C$\to$F scenario and 0.9 otherwise are adopted.   

\begin{table}[t]
\renewcommand{\arraystretch}{0.8}
\caption{
Effect of the amount of shared knowledge from the student model to the teacher model.
}
\label{tab:alpha}

\vspace{-2mm}
\centering
\tabcolsep 10pt
\resizebox{.9\columnwidth}{!}{%
\begin{tabular}{rcccc}
\toprule
& $\alpha$ & C$\to$F & K$\to$C & S$\to$C \\
\toprule
\color{gray} \scriptsize 1 & 0.30 & 23.10 & 42.50 & 51.10 \\
\color{gray} \scriptsize 2 & 0.50 & 26.40 & 46.10 & 53.40 \\
\color{gray} \scriptsize 3 & 0.70 & 30.50 & 48.00 & 54.60 \\
\color{gray} \scriptsize 4 & 0.80 & 33.40 & 52.60 & 57.30 \\
\color{gray} \scriptsize 5 & 0.90 & 36.10 & \textbf{56.80} & \textbf{59.60} \\
\color{gray} \scriptsize 6 & 0.95 & \textbf{37.40} & 56.30 & 57.30 \\
\color{gray} \scriptsize 7 & 0.99 & 35.90 & 55.10 & 57.70 \\
\bottomrule
\end{tabular}
}
\end{table}

\noindent\textbf{Does the teacher matter?} 
As explained before, the teacher model is employed to prevent the student model from collapsing during adaptation. To highlight this, we conduct the adaptation experiments on the three adaptation benchmarks by dropping the teacher model and letting the student learn on its own. We report the mAP at each adaptation iteration as depicted in Fig. \ref{fig:catastrophic}. The effect of the teacher model is evident as the student phases out and fails to maintain a stable trend. In particular, the performance starts from the source-pretrained mAPs reported in Tab.~\ref{tab:c2f} and Tab.
~\ref{tab:k2c_s2c} and flattens after a couple of hundreds of iterations (i.e., note that an adaptation epoch is worth approximately 3000 iterations, yet the performance vanishes in far less than one epoch).  

\begin{figure}
\centering

\begin{tikzpicture}[scale=1]
\begin{axis}[
    title={},
    xlabel={Iterations},
    ylabel={Detection mAP (\%)},
    xmin=0, xmax=500,
    ymin=0, ymax=50,
    xtick={0, 100, 200, 300, 400, 500},
    ytick={0, 10, 20, 30, 40, 50},
    legend pos=north east,
    ymajorgrids=true,
    grid style=dashed,
]

\addplot[
    color=red,
    mark=dot,
    line width=0.7pt
    ]
    coordinates {
(0, 38.6)
(5, 37.6)
(10, 42.8)
(15, 39)
(20, 36.7)
(25, 34.8)
(30, 34.3)
(35, 32.1)
(40, 24.6)
(45, 12.7)
(50, 8.5)
(55, 17.6)
(60, 7.4)
(65, 1.7)
(70, 1.8)
(75, 0.4)
(80, 0)
(85, 0)
(90, 0)
(95, 0.1)
(100, 0.1)
(105, 0.2)
(110, 0.1)
(115, 0.1)
(120, 0)
(125, 0)
(130, 0)
(135, 0)
(140, 0)
(145, 0)
(150, 0)
(155, 0)
(160, 0)
(165, 0)
(170, 0)
(175, 0)
(180, 0)
(185, 0)
(190, 0)
(195, 0)
(200, 0)
(205, 0)
(210, 0)
(215, 0)
(220, 0)
(225, 0)
(230, 0)
(235, 0)
(240, 0)
(245, 0)
(250, 0)
(255, 0)
(260, 0)
(265, 0)
(270, 0)
(275, 0)
(280, 0)
(285, 0)
(290, 0)
(295, 0)
(300, 0)
(305, 0)
(310, 0)
(315, 0)
(320, 0)
(325, 0)
(330, 0)
(335, 0)
(340, 0)
(345, 0)
(350, 0)
(355, 0)
(360, 0)
(365, 0)
(370, 0)
(375, 0)
(380, 0)
(385, 0)
(390, 0)
(395, 0)
(400, 0)
(405, 0)
(410, 0)
(415, 0)
(420, 0)
(425, 0)
(430, 0)
(435, 0)
(440, 0)
(445, 0)
(450, 0)
(455, 0)
(460, 0)
(465, 0)
(470, 0)
(475, 0)
(480, 0)
(485, 0)
(490, 0)
(495, 0)
(500, 0)
    };

\addplot[
    color=green,
    mark=dot,
    line width=0.7pt    
    ]
    coordinates {
(0, 36.7)
(5, 41.7)
(10, 40.4)
(15, 34.8)
(20, 30.5)
(25, 23.3)
(30, 23)
(35, 24.4)
(40, 21.8)
(45, 27)
(50, 31.8)
(55, 32.1)
(60, 29.7)
(65, 23.8)
(70, 23.4)
(75, 23.6)
(80, 24)
(85, 25.6)
(90, 24.3)
(95, 23.9)
(100, 18.9)
(105, 12.2)
(110, 10.8)
(115, 12.6)
(120, 14)
(125, 15.9)
(130, 19.8)
(135, 22.2)
(140, 15.8)
(145, 11.9)
(150, 9.3)
(155, 8.6)
(160, 8.5)
(165, 9.6)
(170, 9.4)
(175, 9.4)
(180, 9.7)
(185, 10.5)
(190, 15.8)
(195, 16.6)
(200, 15)
(205, 14)
(210, 15.6)
(215, 15.7)
(220, 15.3)
(225, 15.3)
(230, 15)
(235, 15.3)
(240, 15.9)
(245, 13.8)
(250, 10)
(255, 4)
(260, 1)
(265, 1)
(270, 1.1)
(275, 1.6)
(280, 0.9)
(285, 0.4)
(290, 0.2)
(295, 0.3)
(300, 0.7)
(305, 1)
(310, 0.4)
(315, 0.3)
(320, 0.2)
(325, 0.1)
(330, 0.1)
(335, 0.1)
(340, 0.1)
(345, 0.2)
(350, 0.1)
(355, 0.1)
(360, 0.1)
(365, 0.1)
(370, 0.1)
(375, 0)
(380, 0)
(385, 0)
(390, 0)
(395, 0)
(400, 0)
(405, 0)
(410, 0)
(415, 0)
(420, 0)
(425, 0)
(430, 0)
(435, 0)
(440, 0)
(445, 0)
(450, 0)
(455, 0)
(460, 0)
(465, 0.1)
(470, 0.1)
(475, 0.2)
(480, 0.3)
(485, 0.4)
(490, 0.4)
(495, 0.3)
(500, 0.3)
    };

\addplot[
    color=blue,
    mark=dot,
    line width=0.7pt    
    ]
    coordinates {
(0, 27.9)
(5, 23)
(10, 14.4)
(15, 10.6)
(20, 10.8)
(25, 11.7)
(30, 11.7)
(35, 9.2)
(40, 10.8)
(45, 10.8)
(50, 9.7)
(55, 9.7)
(60, 9.2)
(65, 9.1)
(70, 9.3)
(75, 8)
(80, 9.3)
(85, 11.1)
(90, 10.1)
(95, 8.9)
(100, 8.2)
(105, 6.8)
(110, 9.8)
(115, 11.8)
(120, 13.4)
(125, 10.8)
(130, 8.2)
(135, 9.4)
(140, 9.9)
(145, 9.9)
(150, 9.7)
(155, 10.3)
(160, 8.8)
(165, 8)
(170, 8.5)
(175, 8.6)
(180, 7.4)
(185, 8.1)
(190, 8.2)
(195, 8)
(200, 6.6)
(205, 8.9)
(210, 9.8)
(215, 12)
(220, 11.8)
(225, 11.4)
(230, 11.3)
(235, 9.5)
(240, 8.3)
(245, 7)
(250, 7.5)
(255, 8.5)
(260, 6.8)
(265, 6.1)
(270, 7.2)
(275, 9.4)
(280, 9.9)
(285, 8.5)
(290, 7.7)
(295, 7.5)
(300, 7.9)
(305, 8.8)
(310, 9.2)
(315, 8)
(320, 8.2)
(325, 7.5)
(330, 7.6)
(335, 7.6)
(340, 7.3)
(345, 6.5)
(350, 6.7)
(355, 7.1)
(360, 5.4)
(365, 3.8)
(370, 3.3)
(375, 2)
(380, 0.2)
(385, 0.2)
(390, 0.1)
(395, 0.2)
(400, 0.7)
(405, 1)
(410, 0.7)
(415, 0.5)
(420, 0.5)
(425, 0.2)
(430, 0.1)
(435, 0)
(440, 0)
(445, 0.6)
(450, 0.6)
(455, 0.3)
(460, 0.3)
(465, 0)
(470, 0)
(475, 0)
(480, 0)
(485, 0)
(490, 0)
(495, 0)
(500, 0)
    };
\legend{
C$\rightarrow$F,
K$\rightarrow$C,
S$\rightarrow$C
}

\end{axis}
\end{tikzpicture}

\vspace{-4mm}
\caption{
Effect of catastrophic forgetting phenomenon when the teacher model is discarded. Best viewed in color.
}
\label{fig:catastrophic}

\end{figure}

\noindent\textbf{Does DACA matter?} 
Earlier, we studied the impact when the teacher model is discarded, causing a premature performance slope.
Here, we study the impact when the DACA part is dropped (i.e., Eq.~\ref{eq_ltot} relaxed to $\ell_{tot} = \ell_s$) leaving only the teacher model to self-adapt.
We report the results in Tab. \ref{tab:nodaca} for all the three scenarios. While in the K$\to$C and the S$\to$C cases, the improvements of introducing DACA amount to 1.50\% and 1.30\% respectively, in the C$\to$F benchmark the gain reaches 3.20\%. This stresses the role of DACA, especially when the domain gap is severe as in C$\to$F. 

\begin{table}[hbt!]
\renewcommand{\arraystretch}{0.8}
\caption{Importance of DACA in the adaptation process.}
\label{tab:nodaca}
\centering
\tabcolsep 10pt

\vspace{-2mm}
\resizebox{\columnwidth}{!}{%
\begin{tabular}{rlrrr}
\toprule
& Scenario & C$\to$F & K$\to$C & S$\to$C \\
\toprule
\color{gray} \scriptsize 1 & Without DACA & 34.20 & 55.30 & 58.30 \\
\color{gray} \scriptsize 2 & With DACA & \textbf{37.40} & \textbf{56.80} & \textbf{59.60} \\
\midrule
\color{gray} \scriptsize 3 & Improvement & \impp{3.20} & \impp{1.50} & \impp{1.30} \\
\bottomrule
\end{tabular}
}
\end{table}

\section{Conclusions}\label{conclusion}

Data augmentation is useful to create informative samples to train object detection models, especially when the annotations are unavailable as in the task of source-free domain adaptation. However, no work has attempted to tailor it so far.
This paper proposed a novel source-free domain adaptation scheme for object detection.
It makes use of data augmentation by mining image regions and their respective confident pseudo-labels and reproducing them into challenging composite images that are leveraged to adapt a detector via self-training.
The method was tested on three adaptation benchmarks, scoring new state-of-the-art on two of them. 

\textbf{Limitations.} Our method augments image regions along with their detections, which may contain false positives that are also augmented, misleading the adaptation of the detector.

\textbf{Future work.} To help eliminate false positives and alleviate their contribution in the adaptation process, we believe that zero shot models \cite{liu2025grounding} can be explored to extract high quality pseudo-labels prior to adaptation with SF-DACA.

\bibliographystyle{plain}
\bibliography{references}

\end{document}